\documentclass[letterpaper]{article} % DO NOT CHANGE THIS
\usepackage[preprint]{aaai2027}
\usepackage[hyphens]{url}  % DO NOT CHANGE THIS
\usepackage{graphicx} % DO NOT CHANGE THIS
\usepackage{natbib}  % DO NOT CHANGE THIS AND DO NOT ADD ANY OPTIONS TO IT
\usepackage{caption} % DO NOT CHANGE THIS AND DO NOT ADD ANY OPTIONS TO IT
\usepackage{amsmath}
\usepackage{amssymb}
\usepackage{booktabs}      % \toprule \midrule \bottomrule \cmidrule
\usepackage[table]{xcolor} % \rowcolor
\usepackage{subcaption}    % \subcaption for (a)/(b) minipage captions
\usepackage{array}         % improved column spec
\usepackage{multirow}
\usepackage{pifont}
\definecolor{deltapos}{RGB}{34,139,34}   % ForestGreen-ish for gains
\definecolor{deltaneg}{RGB}{200,50,50}

\nocopyright

\title{RestoreKV: Recovering Full-Cache Behavior Under Aggressive Query-Agnostic KV Cache Eviction}
\author{
    Changwoo Baek\textsuperscript{\rm 1},
    Seungjun Shin\textsuperscript{\rm 2}\textsuperscript{$\dagger$},
    Kyeongbo Kong\textsuperscript{\rm 1}\textsuperscript{$\dagger$}
}
\affiliations{
    \textsuperscript{\rm 1}Pusan National University\\
    \textsuperscript{\rm 2}Sookmyung Women's University\\
    \{higok18, kbkong\}@pusan.ac.kr, seungjun@sookmyung.ac.kr
}

\begin{document}

\maketitle

\begingroup
\renewcommand\thefootnote{$\dagger$}
\footnotetext{Corresponding authors.}
\endgroup

\begin{abstract}
Query-agnostic KV cache eviction compresses a context once and reuses the
resulting cache for arbitrary future queries, but performance can collapse
under tight budgets. Existing methods primarily improve which original KV
pairs are retained. We introduce RestoreKV, which complements this
selection-based formulation with learned restoration under the same total KV
budget. Our key insight is that, although the information lost through
eviction is context-specific, the mechanism for generating its compact
complement can be shared across contexts. After context prefill, a few restore
tokens attend to the full KV cache in a single LoRA-adapted pass, generating a
compact, context-conditioned restore cache. The base importance scorer and
eviction rule remain unchanged, and the adapters are disabled for all
subsequent queries and decoding. RestoreKV is trained through
parameter-efficient self-distillation from the frozen full-cache model,
optimizing only $0.4\%$ of the parameters and requiring no task-specific
tuning. Across four backbones and four long-context benchmarks, RestoreKV
substantially reduces compression-induced degradation. On Qwen3-4B, it
improves 59 of 60 paired, budget-matched settings across five base eviction
methods; at a $5\%$ budget, it raises KVzip from $38.2$ to $73.2$ on
RULER-4K. Applied to KVzip+, RestoreKV reaches $86.4$ RULER accuracy at
$16\times$ compression on the KVPress Benchmark, while adding less than
$0.5\%$ one-time cache-construction overhead in a 32K-context evaluation.
Our project page is available at \url{https://paper.pnu-cvsp.com/RestoreKV/}.
\end{abstract}

\section{Introduction}

Large language models increasingly rely on long contexts, but their key--value
(KV) caches grow linearly with sequence length. KV cache eviction reduces this
cost by retaining only a subset of cached KV pairs~\citep{
xiao2024efficient,snapkv,h2o,pyramidkv,liu2026chunkkv,
qu2025mobile,feng2026ada,park2026keydiff,
tang2025razorattention}. In particular, query-agnostic eviction methods
compress a context once before future queries are known and reuse the resulting
cache across arbitrary subsequent requests~\citep{kvzip,kvzipplus,contrastkv}.
Recent methods in this setting largely follow the same formulation:
\emph{select a better subset of the original KV pairs}. Although they differ
in their importance signals or learned predictors~\citep{
kvzip,kvzipplus,contrastkv,fastkvzip}, the resulting cache remains composed
only of retained original states. The central question has therefore been
which KV pairs should survive under a fixed budget.

\begin{figure}[t!]
    \centering
    \includegraphics[width=\linewidth]{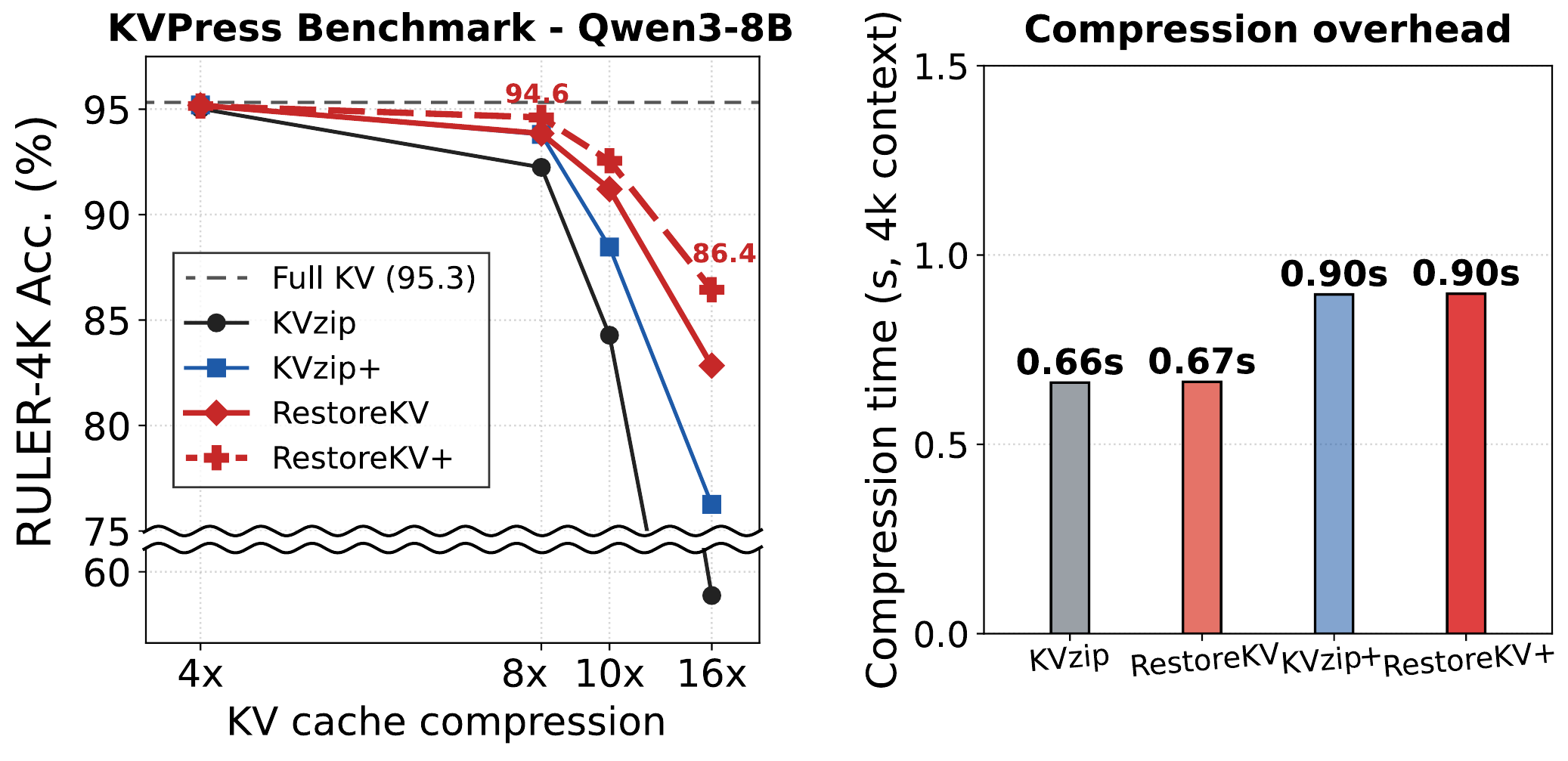}
    \caption{\textbf{RestoreKV narrows the gap to full-cache behavior under
    aggressive compression at negligible one-time cost.}
    Applied to KVzip+, RestoreKV reaches 86.4 RULER accuracy at
    $16\times$ compression on the KVPress Benchmark while adding negligible
    compression-time overhead.}
    \label{fig:teaser}
\end{figure}

\begin{figure*}
    \centering
    \includegraphics[width=\linewidth]{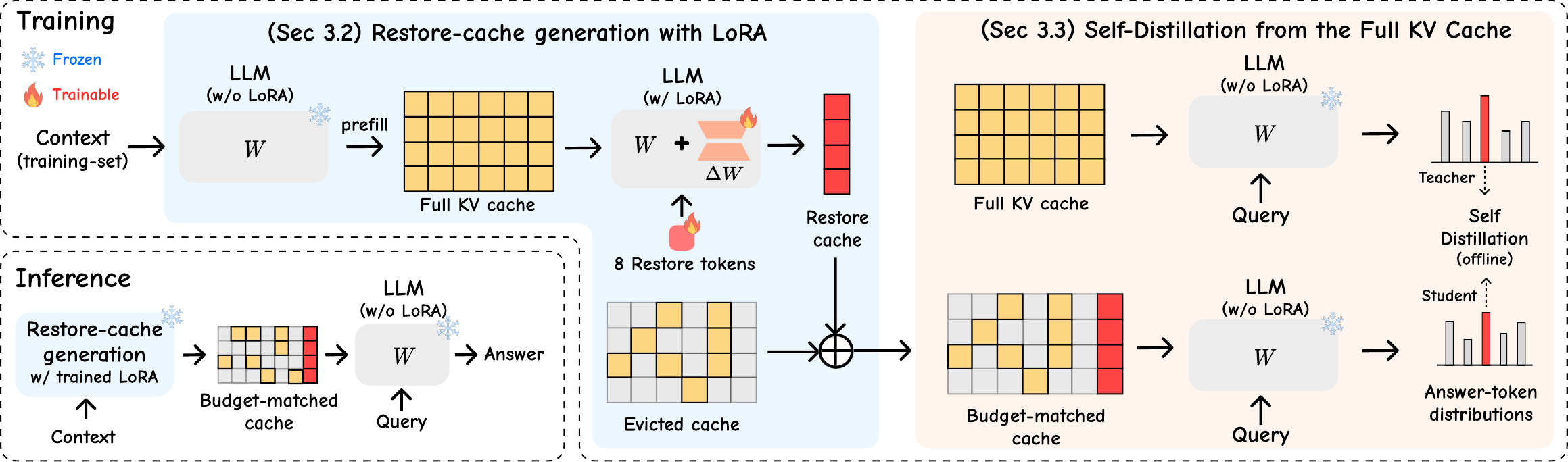}
\caption{\textbf{Overview of RestoreKV.}
After context prefill, a single LoRA-adapted restore pass processes
$n$ restore tokens ($n{=}8$ by default) with causal access to the full KV
cache, generating $nLH$ context-conditioned KV pairs.
These restore states are combined with the retained context states under the
same total KV budget.
LoRA is active only during restore-cache generation; subsequent query
processing and decoding use the original frozen backbone.
\textbf{Training (offline):} The restore-token embeddings and LoRA adapters (0.4\% of the 4B backbone)
are optimized by self-distillation from the frozen full-cache teacher,
\textbf{Inference:} The trained restore tokens and LoRA adapters generate
the restore cache once for each new context.}
    \label{fig:overview}
\end{figure*}

However, better retention alone does not fully prevent degradation under
aggressive eviction. As shown in Fig.~\ref{fig:teaser}, the base eviction
methods remain close to the full-cache model at mild compression but diverge
sharply as the compression ratio increases, whereas RestoreKV substantially
narrows this gap. This motivates a complementary question: rather than relying
only on a better subset, can we generate a small context-conditioned complement
before eviction to preserve behavior that would otherwise be lost? Although
the vulnerable information differs across contexts, the mechanism for
generating its compact complement may be learned across them.

In this work, we introduce \textbf{RestoreKV}, which complements
query-agnostic KV eviction with a context-conditioned restore cache while
preserving the same total KV budget. Our key insight is that, although the
information that would be lost through eviction varies across contexts, the
mechanism for generating its compact complement can be shared across them.
As illustrated in Fig.~\ref{fig:overview}, after context prefill and before
eviction, eight restore tokens attend to the full KV cache in a single
LoRA-adapted restore pass, producing a compact restore cache of
context-conditioned KV pairs. The restore cache occupies a small portion of
the fixed budget, while the base evictor fills the remaining slots with
retained context KV pairs; together, they form the final budget-matched cache.
RestoreKV leaves the base importance scorer and eviction rule unchanged.
Unlike approaches that rewrite selected states or replace the entire prefix
cache, it preserves most retained original states and learns only a small
complement to them. Because the restore pass processes only eight additional
positions once per context and LoRA is disabled afterward, it adds negligible
cache-construction overhead while leaving subsequent query processing and
decoding unchanged.

RestoreKV is trained offline through self-distillation from a frozen
full-cache teacher. For the same context and query, the student using the final
budget-matched cache is optimized to match the teacher's answer-token
distributions. Only the restore-token embeddings and lightweight
LoRA~\citep{hu2022lora} adapters are updated, while the backbone and base
evictor remain frozen. Component ablations clarify what the restore pass
learns: at a $5\%$ KV budget, optimizing the restore embeddings alone improves
KVzip from $38.2$ to $42.1$, whereas fixed embeddings with learned LoRA
adaptation reach $71.9$. Adapting only the $q/k/v$ projections further reaches
$72.4$, close to RestoreKV's $73.2$. These results indicate that the gain
arises primarily from attention-side adaptation that generates a
context-conditioned restore cache, rather than from generic information
memorized in the restore embeddings.

Across four model backbones, four long-context benchmarks, and five base
eviction methods, RestoreKV consistently narrows the gap to full-cache
performance, with larger gains typically observed under tighter budgets. On
Qwen3-4B at a $5\%$ KV budget ratio, it improves KVzip from $38.2$ to $73.2$
on RULER-4K. Applied to KVzip+, RestoreKV reaches $86.4$ RULER accuracy at
$16\times$ compression on the KVPress Benchmark
(Fig.~\ref{fig:teaser}). In our 32K-context evaluation, the eight-token restore
pass adds only $0.03$--$0.04$ seconds, less than $0.5\%$ of the total
compression time, and $84$ MB ($0.4\%$) of peak memory, while preserving the
query-time KV budget and decoding cost of the base method.

Our contributions are summarized as follows:
\begin{itemize}
    \item We introduce a complementary restoration perspective on
    query-agnostic KV eviction: instead of learning only which original KV
    pairs to retain, we learn a shared mechanism that generates a compact,
    context-conditioned restore cache to complement the retained context
    cache.

    \item We propose RestoreKV, a budget-matched, single-pass plug-in that
    preserves the base importance scorer and eviction rule. Its LoRA adapters
    are used only during restore-cache generation, leaving subsequent query
    processing and decoding unchanged.

    \item We demonstrate consistent improvements across five eviction methods,
    four model backbones, and four long-context benchmarks, and show that
    attention-side adaptation is the primary source of recovery under
    aggressive eviction.
\end{itemize}

%%%%%%%%%%%%%%%%%%%%%%%%%%%%%%%%%%%%%%%%%
\section{Related Work}

\subsection{Selection-Based KV Cache Eviction}

Most KV cache eviction methods estimate KV importance using signals associated
with a specific query, such as query-dependent attention or activation
statistics~\citep{snapkv,h2o}. Although effective when the request is already
known, these methods cannot construct a compressed cache before future queries
are observed. Query-agnostic eviction instead compresses a context once and
reuses the resulting cache across arbitrary subsequent requests.
KVzip~\citep{kvzip} estimates the contribution of individual KV pairs through
context reconstruction, while KVzip+~\citep{kvzipplus} and
ContrastKV~\citep{contrastkv} refine the importance signal through output-norm
weighting and contrastive objectives, respectively.

Recent methods further amortize or improve importance estimation using
lightweight trainable modules. KVzap~\citep{kvzipplus} predicts KVzip+
importance scores directly from hidden states, while Fast
KVzip~\citep{fastkvzip} learns sink-attention gates through context
reconstruction. LookaheadKV~\citep{lookaheadkv} considers the query-aware
setting and uses lookahead tokens with selectively activated LoRA modules to
predict response-induced importance without generating a surrogate response.
Despite differences in query availability and scoring mechanisms, these
methods share the same selection-based endpoint: learning determines which
original KV pairs should survive, while the resulting compressed cache remains
a subset of the original cache.

\subsection{Synthesized KV Cache Representations}

Beyond selecting unmodified KV pairs, recent methods expand the
representational space of the compressed cache. KV-Distill~\citep{chari2025kv}
jointly learns an importance scorer and parameter-efficient adapters that
rewrite selected-token representations, using the full-cache predictive
distribution as a distillation target. Cartridges~\citep{eyuboglu2025cartridges}
instead optimizes a context-specific parameterized KV cache through
self-study.

Concurrent work explores several related directions. Attention
Matching~\citep{zweigerfast} constructs compact keys and values by matching
per-head attention outputs and attention mass over reference queries.
VECTOR~\citep{lin2026simple} retains the keys of approximated cache entries
and reconstructs their values during generation using an offline-calibrated
linear map. Latent Context Compilation~\citep{li2026latent} compiles each
context into buffer-token KV states using a disposable LoRA, while
Still~\citep{o2026still} uses learned per-layer compactors to replace the
original prefix cache.

RestoreKV occupies a complementary design point. It keeps the base importance
scorer and eviction rule fixed, preserves most selected original KV pairs
without modification, and reserves only a small portion of the same total
budget for a compact, context-conditioned restore cache. Its key premise is
that the information lost through eviction is context-specific, whereas the
mechanism for generating its compact complement can be shared across contexts.
Trained through full-cache self-distillation, this shared mechanism generates
a restore cache for each new context in a single pre-eviction pass, without
per-context optimization or generation-time reconstruction. Once the
budget-matched cache is constructed, the adapters are disabled, and all
subsequent query processing and decoding use the original frozen model.

\section{Method}
\label{sec:method}

RestoreKV complements an existing query-agnostic KV evictor with a learned
restoration mechanism. As illustrated in Fig.~\ref{fig:overview}, it generates
a compact, context-conditioned restore cache from the full KV cache before
eviction and combines it with the retained context cache under the same total
KV budget. The base importance scorer and eviction rule remain unchanged.
The restoration mechanism is trained through self-distillation from the frozen
full-cache model and is used only during one-time restore-cache construction.

\subsection{Query-Agnostic Eviction Formulation}

Consider a decoder-only Transformer $f_{\theta}$ with $L$ layers and $H$ KV
heads. Given a context $x_{1:T}$, the prefill stage produces a full KV cache
$\mathcal{C}$ containing $TLH$ KV pairs. Generation conditioned on this cache
is denoted by $f_{\theta}(\cdot\mid\mathcal{C})$.

A query-agnostic eviction method assigns importance scores
$\mathbf{s}\in\mathbb{R}^{L\times H\times T}$ without observing future
queries. At KV budget ratio $r$, the total KV budget is
\begin{equation}
    B=\lfloor rTLH\rfloor.
    \label{eq:retained_budget}
\end{equation}
The resulting retained context cache is
\begin{equation}
    \mathcal{C}'
    =
    \operatorname{Evict}
    \left(
        \mathcal{C},
        \mathbf{s},
        B
    \right),
    \label{eq:base_cache}
\end{equation}
where $|\mathcal{C}'|=B$. The operator $\operatorname{Evict}$ follows the
layer- and head-wise allocation rule of the underlying method. Once
constructed, $\mathcal{C}'$ can be reused across arbitrary subsequent queries.

Selection-based eviction constructs $\mathcal{C}'$ solely from original KV
pairs in $\mathcal{C}$. RestoreKV keeps the same importance scores and
allocation rule, but reserves a small portion of the fixed budget for a
generated, context-conditioned restore cache. We next describe its generation,
budget-matched composition, and training objective.

\subsection{Restore-Cache Generation with LoRA}
\label{sec:restore_generation}

RestoreKV introduces $n$ learnable restore-token embeddings
\begin{equation}
    E=[e_1,\ldots,e_n]\in\mathbb{R}^{n\times d},
    \label{eq:restore_embeddings}
\end{equation}
where $d$ is the hidden dimension. After context prefill, the restore tokens
are processed at positions $T+1,\ldots,T+n$ with causal access to the full KV
cache $\mathcal{C}$. A single LoRA-adapted restore pass generates the
context-conditioned restore cache
\begin{equation}
    \mathcal{C}_{\mathrm{res}}
    =
    \operatorname{Restore}_{\theta,\phi}
    \left(E\mid\mathcal{C}\right),
    \label{eq:restore_cache}
\end{equation}
where $\theta$ denotes the frozen backbone parameters and $\phi$ denotes the
LoRA parameters~\citep{hu2022lora}. The restore-token embeddings and LoRA
parameters are shared across contexts, whereas
$\mathcal{C}_{\mathrm{res}}$ is specific to the current context.

For each adapted linear projection $W$, LoRA applies
\begin{equation}
    \Delta W
    =
    \frac{\alpha}{r_{\mathrm{LoRA}}}
    B_{\phi}A_{\phi},
    \qquad
    W_{\mathrm{res}}=W+\Delta W,
    \label{eq:restore_lora}
\end{equation}
where $r_{\mathrm{LoRA}}$ is the LoRA rank, $\alpha$ is the scaling
factor, and $A_{\phi}$ and $B_{\phi}$ are trainable low-rank matrices,
while $W$ remains frozen. LoRA is enabled only during the restore pass;
context prefill and all subsequent query processing and decoding use the
original backbone $f_{\theta}$.

Each restore token produces one KV pair per layer and KV head, yielding
$\lvert\mathcal{C}_{\mathrm{res}}\rvert=nLH$. Because the restore cache is
generated before eviction, it can incorporate information from all context KV
pairs, including those subsequently removed.

\paragraph{Budget matching.}
To preserve the query-time KV budget, RestoreKV reserves $nLH$ of the $B$
cache slots for $\mathcal{C}_{\mathrm{res}}$ and lets the base evictor fill
the remaining $B-nLH$ slots:
\begin{equation}
\begin{aligned}
\widetilde{\mathcal C}
&=
\operatorname{Concat}\!\left(
    \operatorname{Evict}
    \left(\mathcal C,\mathbf s,B-nLH\right),
    \mathcal C_{\mathrm{res}}
\right),\\
\left|\widetilde{\mathcal C}\right|
&=
(B-nLH)+nLH
=
B
=
\left|\mathcal C'\right|.
\end{aligned}
\label{eq:restored_cache}
\end{equation}
Here, $\operatorname{Concat}$ combines the retained context cache and restore
cache without changing their position indices. The base method still
determines the importance scores and layer- and head-wise allocation of the
retained context budget; RestoreKV changes only the cache composition.
Retained context KV pairs preserve their original RoPE~\citep{su2024roformer}
phases, while the restore tokens occupy positions $T+1,\ldots,T+n$ and future
queries begin at position $T+n+1$.

\subsection{Self-Distillation from the Full KV Cache}
\label{sec:self_distillation}

We train the restore-token embeddings $E$ and LoRA parameters $\phi$ through
self-distillation from the frozen full-cache model~\citep{
hinton2015distilling}. Given a training context $x_{1:T}$ and query
$q=(q_1,\ldots,q_Q)$, the full-cache teacher first generates an answer
$y=(y_1,\ldots,y_M)$. The teacher and the restored-cache student then evaluate
the same answer using $\mathcal{C}$ and the final budget-matched cache
$\widetilde{\mathcal{C}}$, respectively:
\begin{equation}
\begin{aligned}
p_i^{\mathrm{full}}(\cdot)
&=
p_{\theta}\!\left(\cdot\mid q,y_{<i},\mathcal{C}\right),\\
p_i^{\mathrm{res}}(\cdot)
&=
p_{\theta}\!\left(\cdot\mid q,y_{<i},
\widetilde{\mathcal{C}}\right).
\end{aligned}
\label{eq:teacher_student_dist}
\end{equation}
The query and teacher answer are used only to define the offline distillation
target; the context-conditioned restore cache is generated without observing
the query, preserving query-agnostic cache construction.

We minimize the token-averaged symmetric KL divergence
\begin{equation}
\mathcal{L}_{\mathrm{distill}}
=
\frac{1}{2M}
\sum_{i=1}^{M}
\left[
\operatorname{KL}\!\left(
p_i^{\mathrm{full}}\middle\|p_i^{\mathrm{res}}
\right)
+
\operatorname{KL}\!\left(
p_i^{\mathrm{res}}\middle\|p_i^{\mathrm{full}}
\right)
\right].
\label{eq:distillation_loss}
\end{equation}
The teacher distributions are detached, and only $E$ and $\phi$ are updated;
the backbone $\theta$ and base importance scorer remain frozen. We uniformly
sample the KV budget ratio,
\begin{equation}
r\sim\mathcal{U}(r_{\min},r_{\max}),
\qquad
0<r_{\min}<r_{\max}<1,
\label{eq:training_ratio}
\end{equation}
so that one shared restoration mechanism supports multiple cache budgets.

\subsection{Inference}
\label{sec:inference}

At inference, RestoreKV generates $\mathcal{C}_{\mathrm{res}}$ once from the
full cache before future queries are observed and constructs the final
budget-matched cache $\widetilde{\mathcal{C}}$ using
Eq.~\eqref{eq:restored_cache}. The same cache can then be reused across
arbitrary subsequent queries. LoRA is enabled only for the $n$-token restore
pass and disabled thereafter, so all query processing and decoding use the
original frozen backbone. Because the restore cache replaces an equal number
of retained context KV pairs, RestoreKV preserves the query-time KV budget and
decoding cost of the base method; its only additional computation is a
one-time forward pass over $n$ positions.

\begin{figure*}[t!]
    \centering
    \includegraphics[width=0.9\linewidth]{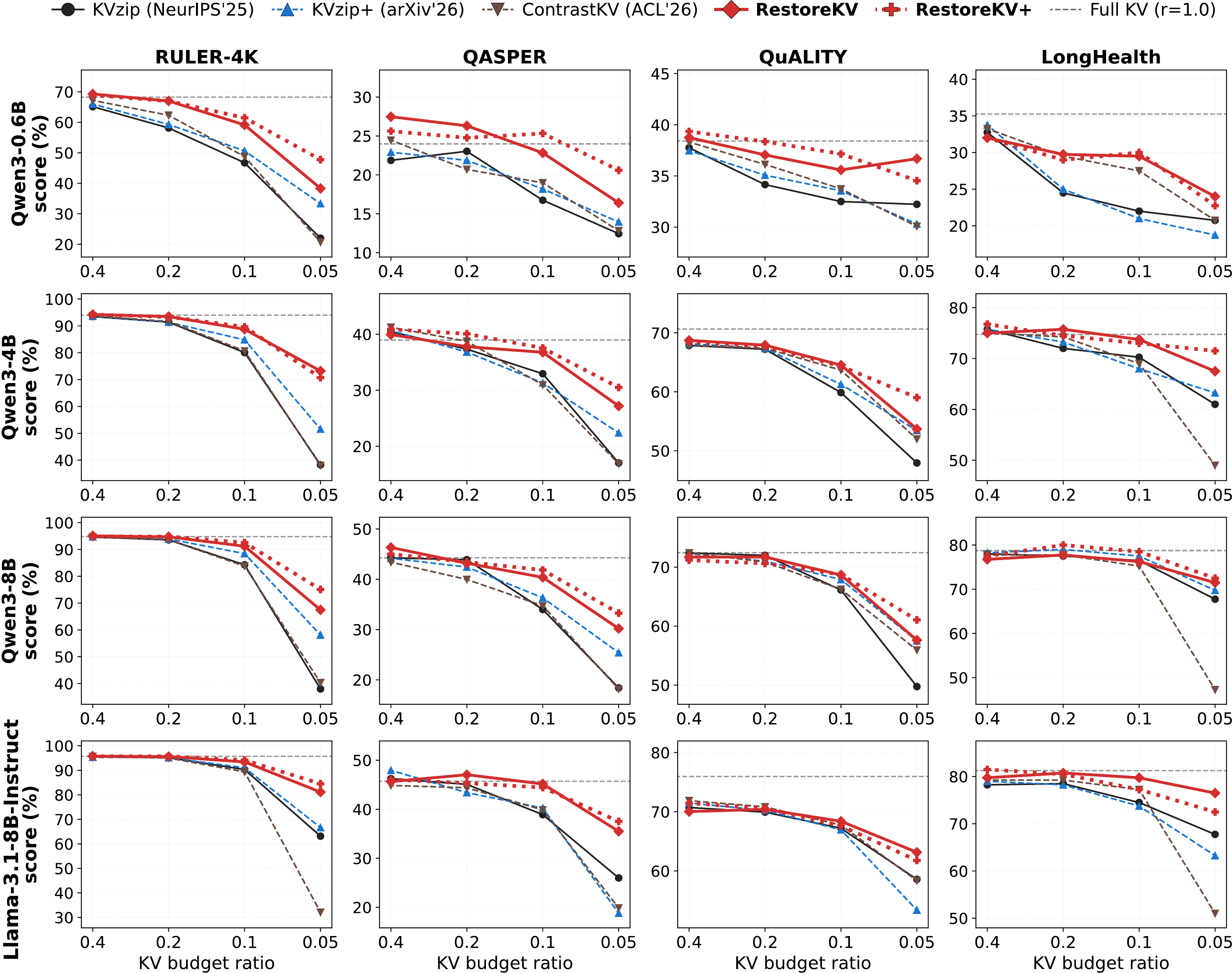}
    \caption{
    \textbf{Performance across four model backbones and four benchmarks.}
    RestoreKV and RestoreKV+ are applied to KVzip and KVzip+, respectively,
    under the same KV budget.
    Both methods reduce compression-induced performance loss, with larger
    gains under tighter cache budgets.
    }
    \label{fig:main}
\end{figure*}

\section{Experiments}
\label{sec:experiments}

We evaluate RestoreKV on four benchmarks under aggressive query-agnostic KV cache eviction. The restore pass is trained \emph{once per model and eviction method}, and
reused across all downstream tasks without any task-specific tuning.

\subsection{Experimental Setup}
\label{sec:experimental_setup}

\paragraph{Datasets.}
We evaluate on four benchmarks spanning distinct query types,
ordered below by increasing context length.
RULER~\citep{ruler} at 4K context is a 13-task synthetic suite covering
retrieval, tracing, and aggregation (fixed 4K); we follow the official
KVPress evaluation protocol~\citep{kvpress}.
QASPER~\citep{qasper}, taken from LongBench~\citep{longbench}, is free-form
QA over scientific papers (avg.\ $\sim$5K tokens).
QuALITY~\citep{quality} is multiple-choice reading comprehension over long
articles ($\sim$6K tokens).
LongHealth~\citep{longhealth} is multiple-choice QA over clinical patient records ($\sim$11K tokens).
Matching the query-agnostic setting, each context is compressed once and
reused across all of its questions.
Additional LongBench results are reported in the appendix.

\paragraph{Models.}
We conduct experiments on Qwen3-0.6B/4B/8B~\citep{qwen3} and
Llama-3.1-8B-Instruct~\citep{llama3}.
The backbone remains frozen in all experiments; only the restore-token
embeddings and LoRA adapters are trained.

\paragraph{Baselines.}
Our primary base methods are KVzip~\citep{kvzip}, KVzip+~\citep{kvzipplus},
and ContrastKV~\citep{contrastkv}, which use reconstruction-based,
output-norm-weighted, and contrastive importance scores, respectively. We
additionally evaluate query-agnostic adaptations of SnapKV~\citep{snapkv} and
H$_2$O~\citep{h2o} to test compatibility with different eviction rules;
implementation details are provided in the appendix.
For every base method, RestoreKV reserves $nLH$ slots from the existing budget
for the restore cache, so each paired comparison uses exactly the same
query-time KV memory. RestoreKV changes neither the base importance scorer nor
its eviction rule. For brevity, figures denote KVzip with RestoreKV as
\textbf{RestoreKV} and KVzip+ with RestoreKV as \textbf{RestoreKV+}; the
``$+$'' refers to the base method. All generations use deterministic greedy
decoding.

\paragraph{Training details.}
We use $n{=}8$ restore tokens with LoRA rank
$r_{\mathrm{LoRA}}{=}8$ and scaling factor $\alpha{=}16$,
corresponding to 0.4\% of the Qwen3-4B backbone parameters.
The training set contains 6.2k context--query pairs derived from
2.5k unique contexts, with some contexts paired with multiple queries.
Contexts $c$ are drawn from LongAlpaca~\citep{chen2024longlora}, PG-19~\citep{pg19}, and
Tulu-3 Flan~\citep{tulu,longpre2023flan}.
Each context is paired with either a source-provided query or one of a fixed set of generic instructions, such as summarizing or explaining the context~\cite{eyuboglu2025cartridges}.
Further details on training-data construction are provided in the appendix.
During training, KV budget ratios are sampled from
$\mathcal{U}(0.025, 0.25)$.
All training and evaluation are conducted on a single NVIDIA RTX PRO 6000
GPU, and a full training run for Qwen3-4B takes approximately two hours.

\begin{table*}[t]
\centering
\small
\setlength{\tabcolsep}{4pt}
\renewcommand{\arraystretch}{0.9}
\resizebox{\textwidth}{!}{%
\begin{tabular}{@{}l ccc ccc ccc ccc@{}}
\toprule
\multirow{2}{*}{Method}
 & \multicolumn{3}{c}{RULER-4K}
 & \multicolumn{3}{c}{QASPER}
 & \multicolumn{3}{c}{QuALITY}
 & \multicolumn{3}{c}{LongHealth} \\
\cmidrule(lr){2-4} \cmidrule(lr){5-7} \cmidrule(lr){8-10} \cmidrule(lr){11-13}
 &  $r{=}0.2$ &  $r{=}0.1$ &  $r{=}0.05$
 &  $r{=}0.2$ &  $r{=}0.1$ &  $r{=}0.05$
 &  $r{=}0.2$ &  $r{=}0.1$ &  $r{=}0.05$
 &  $r{=}0.2$ &  $r{=}0.1$ &  $r{=}0.05$ \\
\midrule
KVzip
& 91.4 & 80.1 & 38.2
& 38.2 & 33.9 & 18.0
& 66.0 & 58.5 & 46.3
& 71.5 & 70.0 & 61.0 \\
\textbf{KVzip + Ours}
& \textbf{93.5}\,{\scriptsize\textcolor{deltapos}{+2.1}}
& \textbf{88.8}\,{\scriptsize\textcolor{deltapos}{+8.7}}
& \textbf{73.2}\,{\scriptsize\textcolor{deltapos}{+35.0}}
& \textbf{39.0}\,{\scriptsize\textcolor{deltapos}{+0.8}}
& \textbf{37.7}\,{\scriptsize\textcolor{deltapos}{+3.8}}
& \textbf{28.7}\,{\scriptsize\textcolor{deltapos}{+10.7}}
& \textbf{67.0}\,{\scriptsize\textcolor{deltapos}{+1.0}}
& \textbf{63.3}\,{\scriptsize\textcolor{deltapos}{+4.8}}
& \textbf{52.4}\,{\scriptsize\textcolor{deltapos}{+6.1}}
& \textbf{75.8}\,{\scriptsize\textcolor{deltapos}{+4.3}}
& \textbf{73.8}\,{\scriptsize\textcolor{deltapos}{+3.8}}
& \textbf{67.5}\,{\scriptsize\textcolor{deltapos}{+6.5}} \\
\midrule
KVzip+
& 91.3 & 84.8 & 51.6
& 38.1 & 32.4 & 23.3
& 66.5 & 60.0 & 51.9
& 73.2 & 67.8 & 63.2 \\
\textbf{KVzip+ + Ours}
& \textbf{93.3}\,{\scriptsize\textcolor{deltapos}{+2.0}}
& \textbf{89.7}\,{\scriptsize\textcolor{deltapos}{+4.9}}
& \textbf{70.7}\,{\scriptsize\textcolor{deltapos}{+19.1}}
& \textbf{40.9}\,{\scriptsize\textcolor{deltapos}{+2.8}}
& \textbf{38.5}\,{\scriptsize\textcolor{deltapos}{+6.1}}
& \textbf{31.6}\,{\scriptsize\textcolor{deltapos}{+8.3}}
& \textbf{66.8}\,{\scriptsize\textcolor{deltapos}{+0.3}}
& \textbf{62.9}\,{\scriptsize\textcolor{deltapos}{+2.9}}
& \textbf{57.9}\,{\scriptsize\textcolor{deltapos}{+6.0}}
& \textbf{74.2}\,{\scriptsize\textcolor{deltapos}{+1.0}}
& \textbf{73.0}\,{\scriptsize\textcolor{deltapos}{+5.2}}
& \textbf{71.5}\,{\scriptsize\textcolor{deltapos}{+8.3}} \\
\midrule
ContrastKV
& 91.6 & 80.7 & 38.0
& 40.1 & 31.9 & 17.8
& 66.2 & 62.2 & 50.7
& 74.2 & 69.0 & 49.0 \\
\textbf{ContrastKV + Ours}
& \textbf{92.3}\,{\scriptsize\textcolor{deltapos}{+0.7}}
& \textbf{84.3}\,{\scriptsize\textcolor{deltapos}{+3.6}}
& \textbf{40.2}\,{\scriptsize\textcolor{deltapos}{+2.2}}
& \textbf{40.9}\,{\scriptsize\textcolor{deltapos}{+0.8}}
& \textbf{36.0}\,{\scriptsize\textcolor{deltapos}{+4.1}}
& \textbf{21.6}\,{\scriptsize\textcolor{deltapos}{+3.8}}
& \textbf{67.6}\,{\scriptsize\textcolor{deltapos}{+1.4}}
& \textbf{63.3}\,{\scriptsize\textcolor{deltapos}{+1.1}}
& \textbf{51.9}\,{\scriptsize\textcolor{deltapos}{+1.2}}
& \textbf{75.2}\,{\scriptsize\textcolor{deltapos}{+1.0}}
& \textbf{70.5}\,{\scriptsize\textcolor{deltapos}{+1.5}}
& \textbf{54.0}\,{\scriptsize\textcolor{deltapos}{+5.0}} \\
\midrule
SnapKV
& 33.8 & 20.6 & 12.7
& 22.4 & 19.0 & 14.3
& 53.9 & \textbf{53.0} & 42.5
& 41.0 & 36.8 & 36.2 \\
\textbf{SnapKV + Ours}
& \textbf{37.7}\,{\scriptsize\textcolor{deltapos}{+3.9}}
& \textbf{26.3}\,{\scriptsize\textcolor{deltapos}{+5.7}}
& \textbf{14.3}\,{\scriptsize\textcolor{deltapos}{+1.6}}
& \textbf{29.9}\,{\scriptsize\textcolor{deltapos}{+7.5}}
& \textbf{24.0}\,{\scriptsize\textcolor{deltapos}{+5.0}}
& \textbf{20.7}\,{\scriptsize\textcolor{deltapos}{+6.4}}
& \textbf{59.4}\,{\scriptsize\textcolor{deltapos}{+5.5}}
& 51.8\,{\scriptsize\textcolor{deltaneg}{$-$1.2}}
& \textbf{49.9}\,{\scriptsize\textcolor{deltapos}{+7.4}}
& \textbf{51.8}\,{\scriptsize\textcolor{deltapos}{+10.8}}
& \textbf{41.8}\,{\scriptsize\textcolor{deltapos}{+5.0}}
& \textbf{38.5}\,{\scriptsize\textcolor{deltapos}{+2.3}} \\
\midrule
H$_2$O
& 8.0 & 3.5 & 3.2
& 22.0 & 13.8 & 14.4
& 59.2 & 50.4 & 41.5
& 60.2 & 48.8 & 37.0 \\
\textbf{H$_2$O + Ours}
& \textbf{17.3}\,{\scriptsize\textcolor{deltapos}{+9.3}}
& \textbf{11.6}\,{\scriptsize\textcolor{deltapos}{+8.1}}
& \textbf{5.7}\,{\scriptsize\textcolor{deltapos}{+2.5}}
& \textbf{30.4}\,{\scriptsize\textcolor{deltapos}{+8.4}}
& \textbf{24.7}\,{\scriptsize\textcolor{deltapos}{+10.9}}
& \textbf{21.3}\,{\scriptsize\textcolor{deltapos}{+6.9}}
& \textbf{60.9}\,{\scriptsize\textcolor{deltapos}{+1.7}}
& \textbf{54.9}\,{\scriptsize\textcolor{deltapos}{+4.5}}
& \textbf{47.7}\,{\scriptsize\textcolor{deltapos}{+6.2}}
& \textbf{64.8}\,{\scriptsize\textcolor{deltapos}{+4.6}}
& \textbf{59.0}\,{\scriptsize\textcolor{deltapos}{+10.2}}
& \textbf{46.8}\,{\scriptsize\textcolor{deltapos}{+9.8}} \\
\bottomrule
\end{tabular}}
\caption{
\textbf{Generalization across KV cache eviction methods.}
RestoreKV improves five different base methods on Qwen3-4B under
matched KV budgets, with larger gains typically observed at tighter budgets.
Colored subscripts indicate absolute changes from the corresponding baseline.
}
\label{tab:method_generalization}
\end{table*}

\subsection{Main Results}
\label{sec:main_results}

Figure~\ref{fig:main} reports performance across all four backbones and
benchmarks, with the full KV cache shown as a dashed reference. Differences
are modest at mild budgets, but the base eviction methods increasingly diverge
from full-cache performance as the budget tightens, whereas RestoreKV remains
substantially closer. On Qwen3-4B at $r{=}0.05$, RestoreKV improves KVzip
from $38.2$ to $73.2$ on RULER-4K, while RestoreKV+ improves KVzip+ from
$51.6$ to $70.7$ under the same total KV budget. Similar trends appear across
the Qwen3 and Llama families and from 0.6B to 8B parameters, showing that the
benefit is not confined to one model scale or architecture.

\paragraph{Generalization across eviction methods.}
Table~\ref{tab:method_generalization} evaluates RestoreKV on five KV cache eviction
methods: KVzip, KVzip+, ContrastKV, and the query-agnostic variants of
SnapKV and H$_2$O following the KVzip setting~\citep{kvzip}.
Although these methods use different importance criteria and eviction
procedures, RestoreKV leaves their scorers and eviction rules unchanged and
simply replaces part of the selected cache with restore cache under the same
total KV budget.
RestoreKV improves the base method in most settings, with larger gains
typically observed under tighter cache budgets.
These results indicate that its effectiveness is not tied to a particular
selection strategy and highlight RestoreKV as a broadly compatible plug-in
for existing KV cache eviction pipelines.

\paragraph{Comparison with a Synthesized-Cache Baseline.}
Attention Matching (AM)~\citep{zweigerfast} constructs a compact cache for
each context by matching per-head attention behavior over reference queries.
In contrast, RestoreKV amortizes restoration across contexts and generates
only a small, context-conditioned complement in a single pre-eviction pass.
Under the same Qwen3-4B setup and matched KV budgets,
RestoreKV+ outperforms AM-fast on RULER-4K and at the tightest LongHealth
budget, while AM-fast is stronger at a milder LongHealth budget.
At $r{=}0.05$, RestoreKV+ achieves $70.7$ versus $52.8$ on RULER-4K and
constructs a 4K-context cache in $0.74$\,s versus $9.68$\,s for AM-fast.
Thus, RestoreKV+ provides competitive or better accuracy at approximately
$13\times$ lower construction time, without per-context query generation or
fitting.

\begin{figure}[t]
    \centering
    \includegraphics[width=\linewidth]{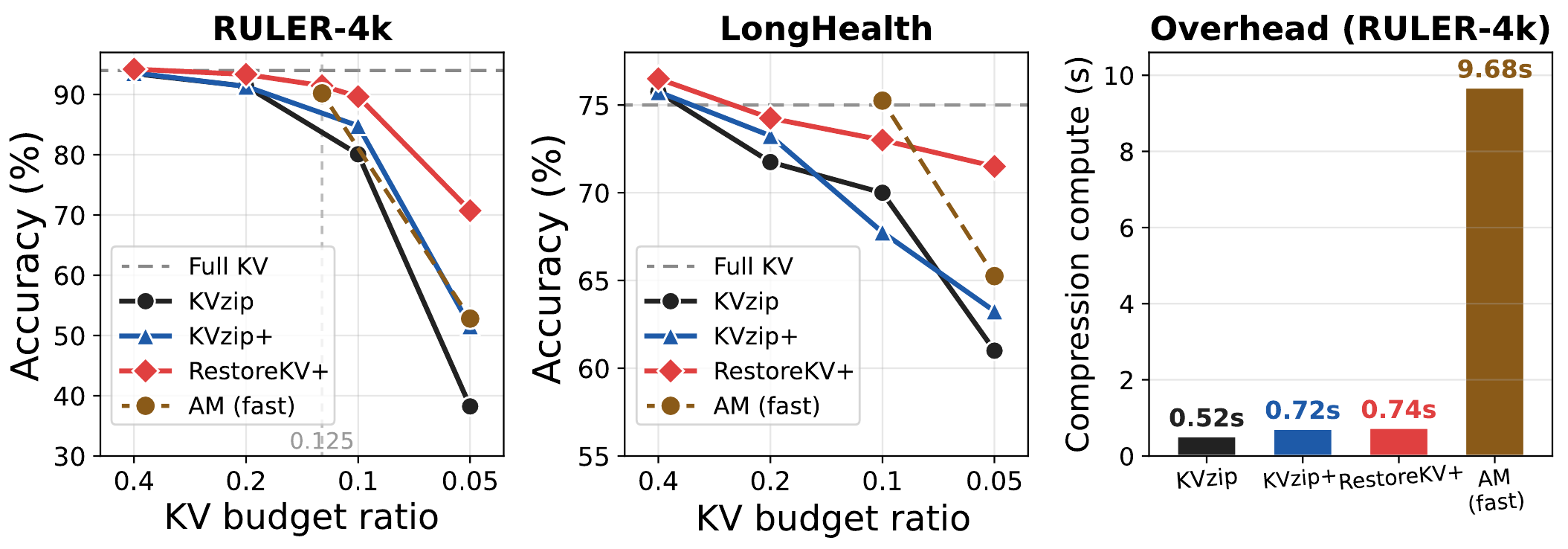}
    \caption{\textbf{Comparison with a Synthesized-Cache Baseline on Qwen3-4B.}
RestoreKV+ consistently improves token-eviction baselines and remains
competitive with AM-fast, a strong synthesized-cache method, at
$\sim$$13\times$ lower compression time on 4K-token contexts.}
    \label{fig:tradeoff}
\end{figure}

\subsection{Analysis of RestoreKV}

\begin{figure}[t]
    \centering
    \includegraphics[width=\linewidth]{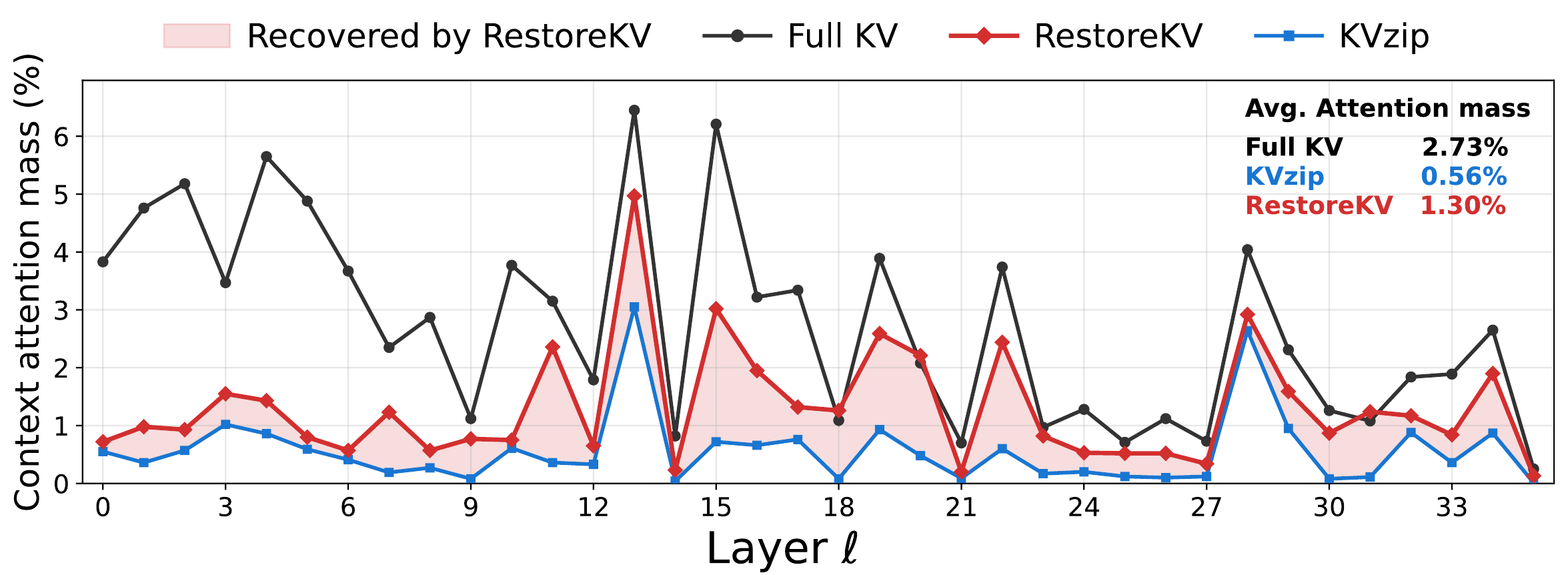}
    \caption{
    \textbf{Layer-wise context-attention recovery on RULER-4K
    (Qwen3-4B, $r{=}0.05$).}
    KVzip substantially suppresses attention from the final query token to
    non-sink context KV pairs. RestoreKV recovers part of the lost attention
    (shaded), increasing the average context-attention mass from $0.56\%$ to
    $1.30\%$ and reducing KL divergence from the full-cache model by
    $16\%$ ($0.198\!\to\!0.167$).}
    \label{fig:mechanism}
\end{figure}

\begin{table}[t]
\centering
\setlength{\tabcolsep}{4pt}
\renewcommand{\arraystretch}{0.95}
\resizebox{\columnwidth}{!}{%
\begin{tabular}{c c c ccc}
\toprule
& & & \multicolumn{3}{c}{KV budget ratio $r$} \\
\cmidrule(lr){4-6}
Restore emb. & LoRA target & \#Params & 0.2 & 0.1 & 0.05 \\
\midrule
-- & -- & -- & 91.4 & 80.1 & 38.2 \\
Learned & -- & 0.02M {\scriptsize(0.0005\%)} & 92.7 & 82.6 & 42.1 \\
Learned & $q/k/v$ & 4.0M {\scriptsize(0.1\%)} & 93.4 & 88.2 & 72.4 \\
Learned & $q/k/v/o$ & 5.9M {\scriptsize(0.15\%)} & 93.5 & 88.4 & 72.0 \\
Fixed & $q/k/v/o{+}$MLP & 16.5M {\scriptsize(0.4\%)} & 93.5 & 88.0 & 71.9 \\
Learned & $q/k/v/o{+}$MLP & 16.5M {\scriptsize(0.4\%)} & \textbf{93.5} & \textbf{88.8} & \textbf{73.2} \\
\bottomrule
\end{tabular}}
\caption{\textbf{Component and LoRA-target ablations on RULER-4K
(Qwen3-4B).}
The first row is KVzip without a restore cache; the last row is the full
RestoreKV configuration.
``Fixed'' replaces all eight learned restore-token embeddings with the
embedding of the line-break token \texttt{\textbackslash n}.}
\label{tab:abl_restore_components}
\end{table}

\paragraph{Context-attention recovery.}
Figure~\ref{fig:mechanism} compares the layer-wise attention assigned by the
final query token to non-sink context KV pairs. Under aggressive eviction,
KVzip suppresses context attention across most layers and misses many of the
high-attention peaks observed with the full cache, reducing the average mass
from $2.73\%$ to $0.56\%$ and leaving $7$ of $36$ layers below $0.1\%$.
RestoreKV recovers part of this lost attention across layers, as highlighted
by the shaded region, increasing the average mass to $1.30\%$ and raising
every layer above $0.1\%$. Its layer-wise profile also more closely follows
the full-cache pattern, reducing KL divergence from $0.198$ to
$0.167$. Consistent with this recovery, the divergence of the final predictive
distribution from the full-cache model decreases from $7.3$ to $3.8$.
These results indicate that RestoreKV partially restores both the amount and
the layer-wise allocation of context attention, accompanied by predictions
closer to those of the full-cache model.

\paragraph{Source of restoration gains.}
Table~\ref{tab:abl_restore_components} separates the roles of the
restore-token embeddings and LoRA adaptation. All restore variants use
$n{=}8$ and the same total KV budget. At $r{=}0.05$, learning only the
restore embeddings reaches $42.1$, whereas fixed embeddings with LoRA reach
$71.9$, recovering $96\%$ of RestoreKV's full improvement to $73.2$.
Restricting LoRA to the $q/k/v$ projections achieves $72.4$ with only
$4.0$M trainable parameters. These results indicate that restoration is
driven primarily by attention-side adaptation that generates the
context-conditioned restore cache, while learned embeddings and broader
backbone adaptation provide only marginal additional gains.

\subsection{Ablations}
\label{sec:ablation}

\paragraph{Effect of full-context conditioning.}
Table~\ref{tab:abl_prunefirst} compares generating the restore cache before
eviction with generating it from the already evicted cache.
At $r{=}0.05$, the evicted-cache variant improves KVzip from $38.2$ to
$64.4$, indicating that much of the recovery comes from the learned restore
transformation rather than full-context access alone.
Conditioning on the full cache further improves accuracy to $73.2$, providing
an additional $8.8$ points.

Because the two variants use the same restore tokens, adaptation, KV budget,
and positional offset, this gap isolates the benefit of full-context
conditioning.
Together with Table~\ref{tab:abl_restore_components}, these results suggest
that effective restoration is driven primarily by the learned attention-side
transformation, while access to the complete context before eviction provides
a substantial complementary benefit.

\begin{table}[t]
\centering
\small
\setlength{\tabcolsep}{4pt}
\renewcommand{\arraystretch}{0.8}
\begin{tabular}{lccc}
\toprule
& \multicolumn{3}{c}{KV budget ratio $r$} \\
\cmidrule(lr){2-4}
Method & 0.2 & 0.1 & 0.05 \\
\midrule
KVzip ($n{=}0$)             & 91.4 & 80.1 & 38.2 \\
Generated from evicted cache          & \textbf{93.7} & 87.6 & 64.4 \\
\textbf{Generated from full KV cache} & 93.5 & \textbf{88.8} & \textbf{73.2} \\
\bottomrule
\end{tabular}
\caption{\textbf{Effect of full-context conditioning on RULER-4K with Qwen3-4B.}
Generating the restore cache from the full cache before eviction adds
$8.8$ points at $r{=}0.05$ over generating it from the already evicted
cache.}
\label{tab:abl_prunefirst}
\end{table}

\begin{table}[t]
\centering
\small
\setlength{\tabcolsep}{4pt}
\renewcommand{\arraystretch}{0.8}
\begin{tabular}{lccc}
\toprule
& \multicolumn{3}{c}{KV budget ratio $r$} \\
\cmidrule(lr){2-4}
Method & 0.2 & 0.1 & 0.05 \\
\midrule
$n=0$ (KVzip)             & 91.4 & 80.1 & 38.2 \\
$n=1$                     & 93.0 & 85.3 & 65.3 \\
$n=2$                     & 93.4 & 87.6 & 70.4 \\
$n=4$                     & 93.4 & 87.0 & 68.5 \\
$n=8$ \textbf{(default)}  & \textbf{93.5} & \textbf{88.8} & \textbf{73.2} \\
$n=16$                    & 93.3 & 87.8 & 69.1 \\
\bottomrule
\end{tabular}
\caption{\textbf{Ablation on the number of restore tokens (RULER-4K, Qwen3-4B).}
All variants share the same training recipe and total KV budget.}
\label{tab:abl_ntok}
\end{table}

\paragraph{Number of restore tokens.}
Table~\ref{tab:abl_ntok} examines the effect of the number of
restore tokens $n$.
Even a single restore token yields a substantial improvement under aggressive
eviction, increasing accuracy from $38.2$ to $65.3$ at $r{=}0.05$.
Despite minor fluctuations, performance generally improves as more restore
tokens are added up to $n{=}8$, which achieves the best accuracy of $73.2$.
All configurations with restore tokens consistently outperform the
no-restore baseline, while increasing $n$ beyond eight provides no further
gain.
We therefore use $n{=}8$ as the default configuration.

\begin{figure}[t]
    \centering
    \includegraphics[width=0.9\linewidth]{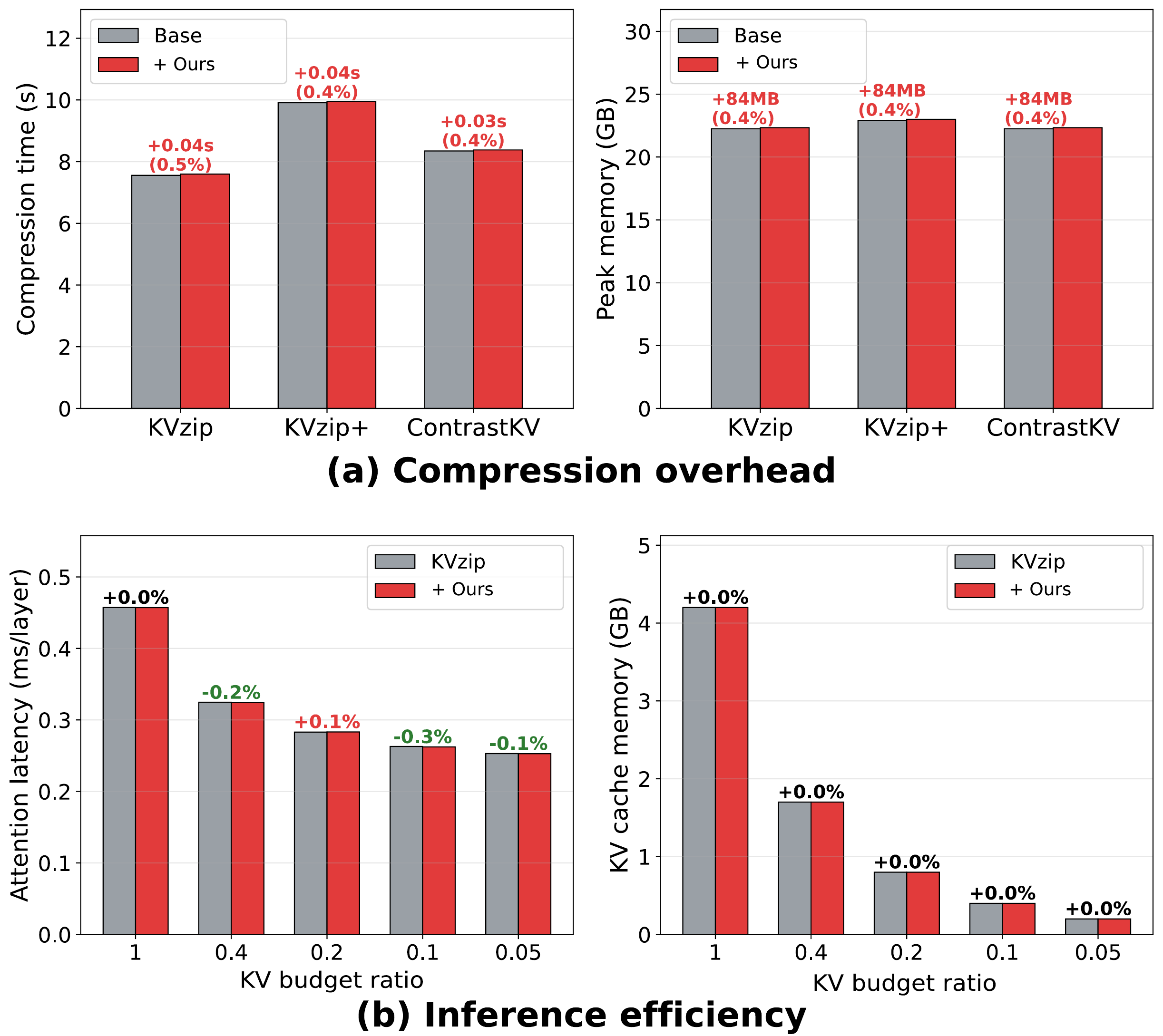}
    \caption{\textbf{Efficiency analysis (Llama-3.1-8B-Instruct, 32K context).}
(a) Restore-cache generation adds only ${\sim}0.04$\,s ($0.5\%$) and $84$\,MB ($0.4\%$) on top of any base scorer.
(b) Attention latency and KV memory are nearly identical to the base method.}
\label{fig:efficiency}
\end{figure}

\subsection{Efficiency Analysis}
Figure~\ref{fig:efficiency} evaluates RestoreKV on Llama-3.1-8B-Instruct with
32K-token contexts using an NVIDIA RTX PRO 6000. Constructing the eight-token
restore cache before eviction incurs only
$0.03$--$0.04$s of additional latency, accounting for less than
$0.5\%$ of the end-to-end compression time, including prefill. It also
increases peak memory during prefill by only $84$\,MB ($0.4\%$),
due to the resident LoRA weights, whose memory cost is constant with respect
to context length.
This small overhead arises because the restore pass is a single forward pass
over eight additional positions after the context has already been prefilled.
During inference, RestoreKV replaces an equal number of context KV pairs
and therefore preserves the same total cache budget.
Across KV budget ratios from $1.0$ to $0.05$, its per-layer attention
latency is nearly identical to KVzip
($0.457$--$0.253$\,ms), while KV-cache memory
is exactly matched, decreasing from $4.20$ to $0.20$\,GB for both methods.
Thus, RestoreKV introduces negligible one-time compression overhead and
no measurable additional inference-time cost.

\section{Conclusion}
\label{sec:conclusion}

We revisited query-agnostic KV cache eviction from a complementary restoration
perspective. RestoreKV uses a shared, LoRA-adapted restoration mechanism to
generate a compact, context-conditioned restore cache before eviction and
combines it with retained original KV pairs under the same total budget. It
preserves the base importance scorer and eviction rule, and disables the
adapters after cache construction, adding negligible one-time overhead and no
query-time KV-memory or decoding cost. Across four backbones, four benchmarks, and five base eviction methods, RestoreKV consistently reduces compression-induced degradation, with its largest gains under aggressive budgets. Our analyses show
that attention-side adaptation is the primary source of recovery, while
full-context conditioning provides an additional complementary benefit.

\bibliography{aaai2027}

\clearpage
\appendix
\setcounter{secnumdepth}{2}

\setcounter{figure}{0}
\setcounter{table}{0}
\renewcommand{\thefigure}{\Alph{figure}}
\renewcommand{\thetable}{\Alph{table}}

\section*{Appendix Overview}
This appendix provides additional experimental results, control experiments,
and implementation details supporting the main paper.

\begin{itemize}
\setlength{\itemsep}{2pt}
\setlength{\parsep}{0pt}
\setlength{\topsep}{2pt}
\setlength{\partopsep}{0pt}

\item \textbf{\ref{app:additional_results}.~Additional Experimental Results}\\
\hspace*{1.5em}\ref{app:fastkvzip}~Applicability to a Learning-Based Scorer\\
\hspace*{1.5em}\ref{app:position_shift}~Effect of the Query-Position Offset\\
\hspace*{1.5em}\ref{app:cross}~Effect of Training–Inference Evictor Mismatch\\
\hspace*{1.5em}\ref{app:training_ratio_range}~Effect of the Training Ratio Range\\
% \hspace*{1.5em}\ref{app:vector_comparison}~Comparison with Other Plug-ins for KV Cache Eviction\\
\item \textbf{\ref{app:bench}~Additional Benchmark Results}\\
% \label{app:bench}
\hspace*{1.5em}\ref{longbench}~LongBench Results\\
\hspace*{1.5em}\ref{scbench}~SCBench Results\\
\item \textbf{\ref{app:implementation}~Implementation and Reproducibility Details}\\
\hspace*{1.5em}\ref{app:traindata}~Training Data Construction\\
\hspace*{1.5em}\ref{app:training_setup}~Training Setup Details\\
\hspace*{1.5em}\ref{app:baseline_adaptations}~Baseline Adaptations (SnapKV and H$_2$O)\\
\hspace*{1.5em}\ref{app:environment}~Experimental Environment\\
\hspace*{1.5em}\ref{app:training_robustness}~Sensitivity to Training Seeds\\
\hspace*{1.5em}\ref{app:eval_metrics}~Evaluation Metrics
\end{itemize}

\section{Additional Experimental Results}
\label{app:additional_results}

\subsection{Applicability to a Learning-Based Scorer}
\label{app:fastkvzip}

\begin{figure}[t]
    \centering
    \includegraphics[width=0.65\linewidth]{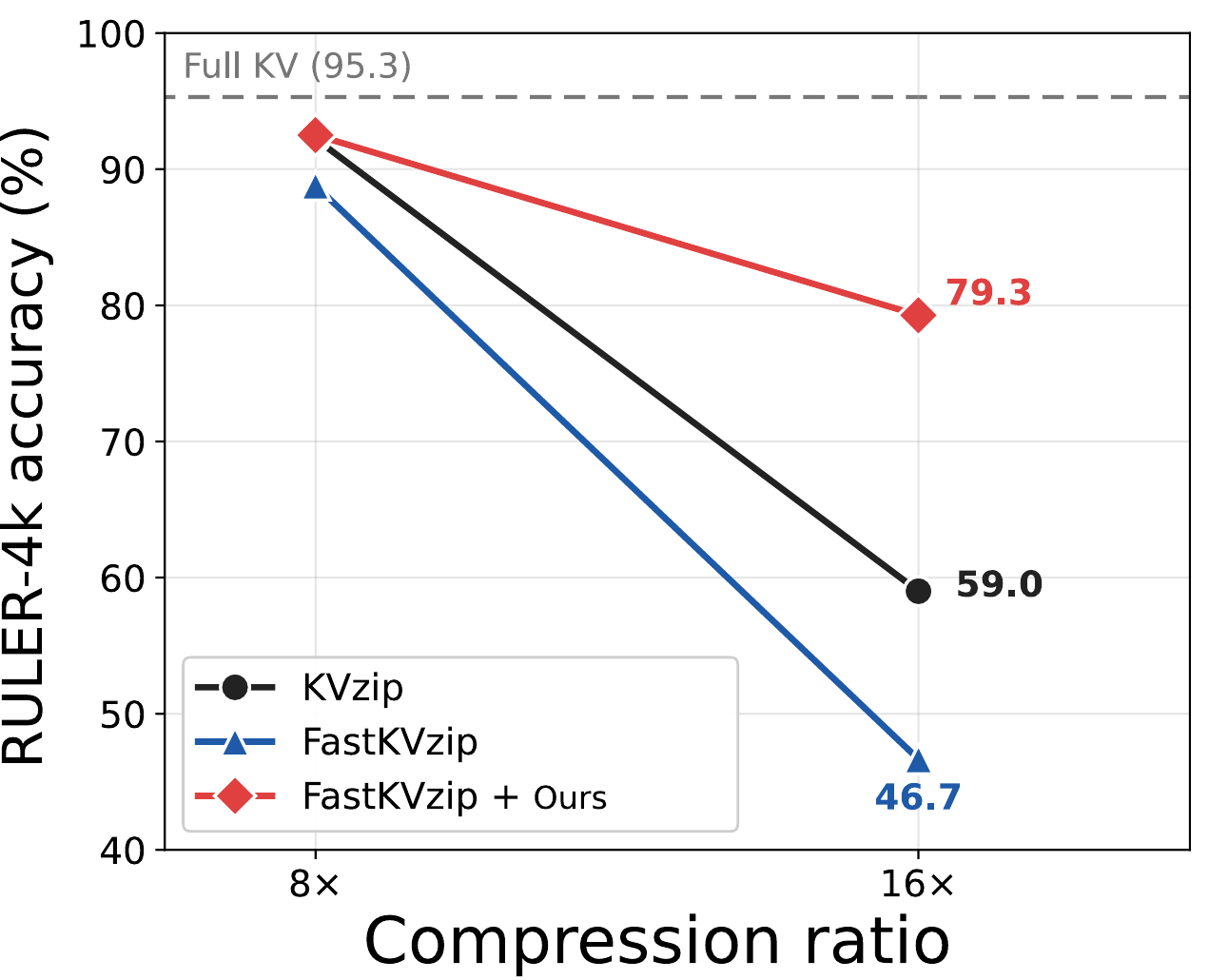}
\caption{\textbf{RestoreKV composes with Fast KVzip, a learning-based scorer}
(RULER-4K, Qwen3-8B). At $16\times$, Fast KVzip drops to $46.7$, RestoreKV
restores it to $79.3$, above plain KVzip ($59.0$).}
    \label{fig:a}
\end{figure}

RestoreKV is also compatible with learned importance scorers. Fast
KVzip~\citep{fastkvzip} replaces KVzip's expensive scoring pass with
lightweight per-layer gates that predict importance from hidden states in a single forward pass. Attaching RestoreKV raises its RULER-4K~\cite{ruler} accuracy from
$46.7$ to $79.3$ at $16\times$ compression (Fig.~\ref{fig:a}), a
$32.6$-point gain that also exceeds standard KVzip ($59.0$). Because Fast KVzip retains single-pass scoring, this result shows that RestoreKV can substantially improve a learned scorer without sacrificing its primary efficiency advantage.

\subsection{Effect of the Query-Position Offset}
\label{app:position_shift}

RestoreKV appends $8$ restore tokens after the context, causing subsequent
queries to begin $8$ positions later than in standard KVzip.
This changes the relative RoPE~\citep{su2024roformer} offsets between the query and the retained
context KV pairs, raising the possibility that the position shift itself
contributes to the observed improvement.
To isolate this effect, we evaluate a control variant of KVzip that uses no
restore tokens or restore cache but shifts the query positions by the same
$+8$ offset.
As shown in Table~\ref{tab:posshift}, this variant achieves $36.9$ at
$r{=}0.05$, comparable to standard KVzip at $38.2$ and substantially below
RestoreKV at $73.2$.
These results indicate that the performance gain cannot be explained by the
RoPE offset alone and instead support the contribution of the
context-conditioned restore cache.

\begin{table}[t!]
\centering
\small
\setlength{\tabcolsep}{5pt}
\renewcommand{\arraystretch}{0.8}
\begin{tabular}{cc cccc}
\toprule
\multicolumn{2}{c}{Training ratio range}
& \multicolumn{4}{c}{KV budget ratio $r$} \\
\cmidrule(lr){1-2}
\cmidrule(lr){3-6}
$r_{\min}$ & $r_{\max}$
& $0.4$ & $0.2$ & $0.1$ & $0.05$ \\
\midrule
 - & -
& 93.5 & 91.4 & 80.1 & 38.2 \\
$0.025$ & $0.25$
& \textbf{94.3} & \textbf{93.5} & \textbf{88.8} & \textbf{73.2} \\
$0.025$ & $0.50$
& 94.1 & 93.1 & 86.1 & 63.5 \\
$0.050$ & $0.25$
& \textbf{94.3} & 93.2 & 86.5 & 65.6 \\
$0.050$ & $0.50$
& 94.0 & 92.8 & 84.5 & 51.5 \\
\bottomrule
\end{tabular}
\caption{\textbf{Ablation on the training ratio range
(RULER-4K, Qwen3-4B).}
RestoreKV is trained by sampling the retention ratio from
$\mathcal{U}(r_{\min}, r_{\max})$.
The first row reports the KVzip baseline without restoration.}
\label{tab:training_ratio_range}
\end{table}

\begin{table}[t!]
\centering
\footnotesize
\setlength{\tabcolsep}{2.7pt}
\renewcommand{\arraystretch}{0.8}
\begin{tabular}{@{}lccccc@{}}
\toprule
\multirow{2}{*}{Variant}
& \multirow{2}{*}{\shortstack{Restore\\states}}
& \multirow{2}{*}{\shortstack{Query\\offset}}
& \multicolumn{3}{c}{KV budget ratio $r$} \\
\cmidrule(lr){4-6}
& & & 0.2 & 0.1 & 0.05 \\
\midrule
KVzip
& No & 0
& 91.4 & 80.1 & 38.2 \\

Offset-only KVzip
& No & $+8$
& 90.5 & 80.8 & 36.9 \\

\textbf{RestoreKV}
& Yes & $+8$
& \textbf{93.5} & \textbf{88.8} & \textbf{73.2} \\
\bottomrule
\end{tabular}
\caption{\textbf{Effect of the query-position offset}
(RULER-4K, Qwen3-4B).
Offset-only KVzip applies the same $+8$ query offset as RestoreKV without
adding restore states. Its performance remains close to standard KVzip,
suggesting that the positional offset alone is unlikely to account for
RestoreKV's improvement.}
\label{tab:posshift}
\end{table}

\begin{table}[t!]
\centering
\small
\setlength{\tabcolsep}{6pt}
\renewcommand{\arraystretch}{1.0}
\begin{tabular}{@{}ccccc@{}}
\toprule
\multirow{2}{*}{Training evictor}
& \multirow{2}{*}{Inference evictor}
& \multicolumn{3}{c}{KV budget ratio $r$} \\
\cmidrule(lr){3-5}
& & $0.20$ & $0.10$ & $0.05$ \\
\midrule
-
& KVzip
& 91.4
& 80.1
& 38.2 \\
SnapKV
& KVzip
& \textbf{93.5}
& 81.6
& 42.0 \\
KVzip
& KVzip
& \textbf{93.5}
& \textbf{88.8}
& \textbf{73.2} \\
\bottomrule
\end{tabular}
\caption{\textbf{Effect of training--inference evictor mismatch
(RULER-4K, Qwen3-4B).}
The inference evictor is fixed to KVzip, while RestoreKV is trained using the
evictor in the first column.}
\label{tab:cross_evictor_transfer}
\end{table}

\begin{table*}[t!]
\centering
\small
\setlength{\tabcolsep}{4pt}
\renewcommand{\arraystretch}{1.15}

% ================= Llama-3.1-8B-Instruct =================
\begin{minipage}{0.5\textwidth}
\centering
\resizebox{\linewidth}{!}{%
\begin{tabular}{lcccccccc}
\toprule
\textbf{Method}
& \textbf{S-QA}
& \textbf{M-QA}
& \textbf{Summ}
& \textbf{Fewshot}
& \textbf{Synth}
& \textbf{Code}
& \textbf{Avg}
& \textbf{Rel.} \\
\midrule

\rowcolor{gray!8}
\multicolumn{9}{c}{\emph{Full KV cache}} \\
Llama-3.1-8B-Instruct
& 44.6 & 47.6 & 29.2 & 54.0 & 55.1 & 48.1 & 46.4 & 100.0\% \\

\midrule
\rowcolor{gray!8}
\multicolumn{9}{c}{\emph{KV budget ratio = 0.125}} \\
KVzip
& 42.7
& \textbf{43.9}
& 27.7
& 61.9
& 44.0
& 47.7
& 44.7
& 96.3\% \\

\textbf{KVzip + Ours}
& \textbf{43.6}
& 43.0
& \textbf{28.5}
& \textbf{63.9}
& \textbf{45.9}
& \textbf{49.3}
& \textbf{45.7}
& \textbf{98.4\%} \\

\midrule
\rowcolor{gray!8}
\multicolumn{9}{c}{\emph{KV budget ratio = 0.0625}} \\
KVzip
& 33.3
& 32.4
& 24.0
& 49.1
& 20.6
& 41.6
& 33.5
& 72.1\% \\

\textbf{KVzip + Ours}
& \textbf{38.2}
& \textbf{35.4}
& \textbf{27.0}
& \textbf{59.3}
& \textbf{22.8}
& \textbf{43.2}
& \textbf{37.7}
& \textbf{81.2\%} \\

\bottomrule
\end{tabular}%
}
\subcaption{Llama-3.1-8B-Instruct}
\end{minipage}
\hfill
% ================= Qwen3-8B =================
\begin{minipage}{0.48\textwidth}
\centering
\resizebox{\linewidth}{!}{%
\begin{tabular}{lcccccccc}
\toprule
\textbf{Method}
& \textbf{S-QA}
& \textbf{M-QA}
& \textbf{Summ}
& \textbf{Fewshot}
& \textbf{Synth}
& \textbf{Code}
& \textbf{Avg}
& \textbf{Rel.} \\
\midrule

\rowcolor{gray!8}\multicolumn{9}{c}{\emph{Full KV cache}} \\
Qwen3-8B & 42.8 & 48.6 & 27.6 & 57.4 & 50.5 & 62.6 & 48.2 & 100.0\% \\
\midrule
\rowcolor{gray!8}\multicolumn{9}{c}{\emph{KV budget ratio = 0.125}} \\
KVzip & 38.4 & \textbf{44.0} & 26.2 & 63.1 & \textbf{42.4} & 56.8 & 45.2 & 93.6\% \\
\textbf{KVzip + Ours} & \textbf{39.2} & 42.6 & \textbf{26.9} & \textbf{64.1} & 39.9 & \textbf{59.9} & \textbf{45.4} & \textbf{94.2\%} \\
\midrule
\rowcolor{gray!8}\multicolumn{9}{c}{\emph{KV budget ratio = 0.0625}} \\
KVzip & 23.2 & 15.5 & 20.9 & 50.8 & 14.3 & 46.9 & 28.6 & 59.3\% \\
\textbf{KVzip + Ours} & \textbf{28.1} & \textbf{17.7} & \textbf{23.8} & \textbf{55.0} & \textbf{15.0} & \textbf{53.0} & \textbf{32.1} & \textbf{66.5\%} \\

\bottomrule
\end{tabular}%
}
\subcaption{Qwen3-8B}
\end{minipage}

\caption{\textbf{LongBench results over 16 tasks under aggressive KV compression.} Categories are single-document QA (S-QA), multi-document QA (M-QA), summarization (Summ), few-shot learning (Fewshot), synthetic tasks (Synth), and code tasks (Code). \textit{Rel.} denotes the average score relative to the full KV cache.}
\label{tab:longbench_restorekv}
\end{table*}

\begin{figure}
    \centering
    \includegraphics[width=0.65\linewidth]{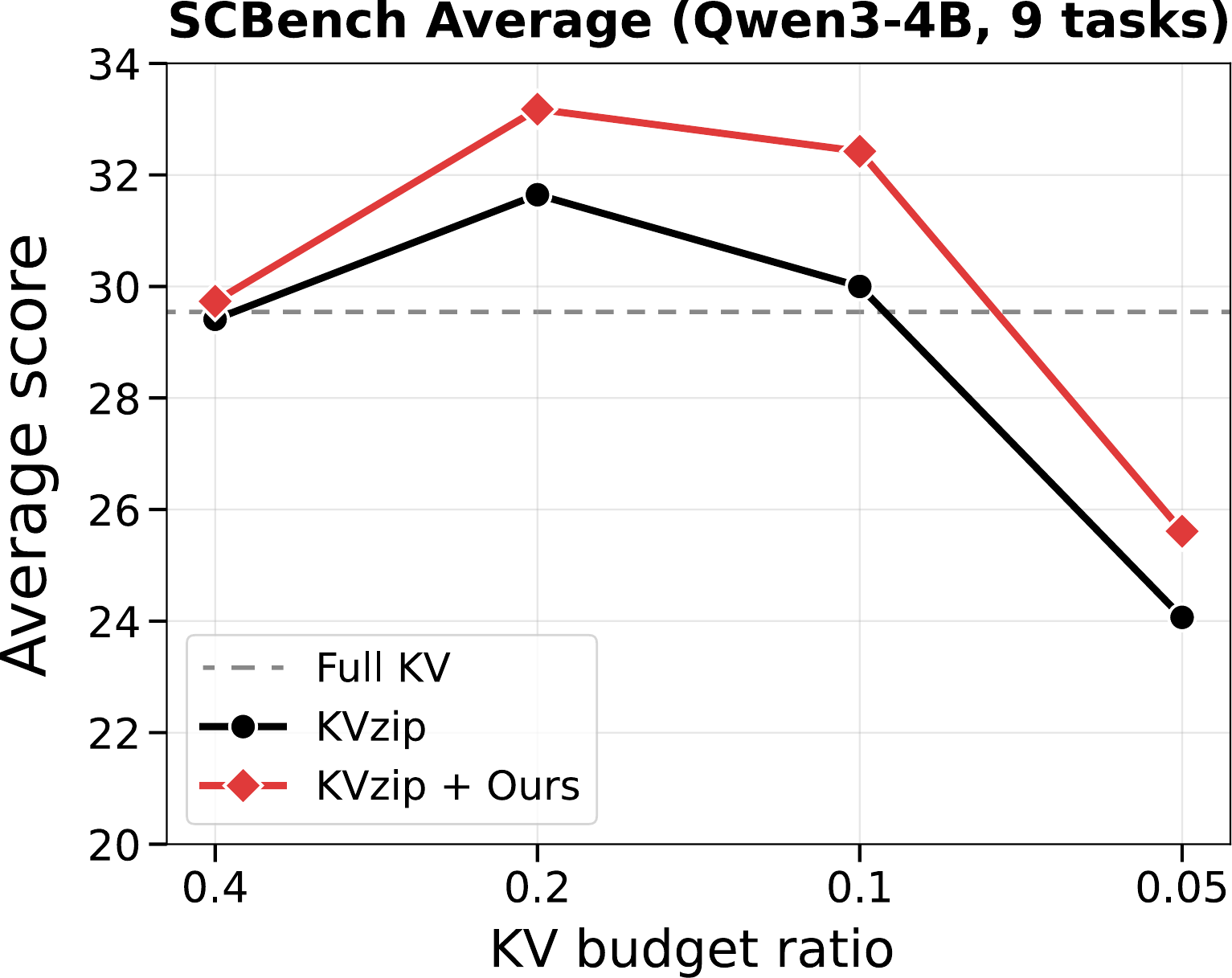}
    \caption{\textbf{SCBench results on long-context tasks} (Qwen3-4B).}
    \label{fig:scbench}
\end{figure}

\subsection{Effect of Training--Inference Evictor Mismatch}
\label{app:cross}

We examine whether a RestoreKV checkpoint trained with one base evictor
remains effective when paired with another at inference. We fix the inference
evictor to KVzip and compare checkpoints trained using either KVzip or SnapKV~\citep{snapkv}
(Table~\ref{tab:cross_evictor_transfer}). At a mild budget
($r{=}0.20$), the matched and mismatched checkpoints perform similarly.
However, matching becomes increasingly important as eviction becomes more
aggressive. At $r{=}0.10$, the SnapKV-trained checkpoint reaches $81.6$,
compared with $88.8$ for the KVzip-trained checkpoint; at $r{=}0.05$, it
reaches $42.0$, compared with $73.2$. The mismatched checkpoint still
slightly outperforms the no-restoration KVzip baseline at the tightest budget
($42.0$ vs.\ $38.2$), indicating limited cross-evictor transfer.
These results suggest that the restoration mechanism adapts to the compression
behavior induced by the training-time evictor.

\subsection{Effect of the Training Ratio Range}
\label{app:training_ratio_range}

Table~\ref{tab:training_ratio_range} examines the effect of the training
range used to sample the KV budget ratio,
$r \sim \mathcal{U}(r_{\min}, r_{\max})$, on RULER-4K.
All configurations achieve comparable performance at the mild evaluation
budget of $r{=}0.4$.
However, the differences become increasingly pronounced as the budget
tightens.
The default range, $\mathcal{U}(0.025, 0.25)$, achieves the strongest
overall performance, reaching 88.8 at $r{=}0.1$ and 73.2 at $r{=}0.05$.
Increasing either the lower bound or the upper bound reduces performance
under aggressive eviction.
These results indicate that training with sufficient exposure to tight KV
budgets is important for effective restoration under severe compression.
We therefore use $\mathcal{U}(0.025, 0.25)$ as the default training range.

\section{Additional Benchmark Results}
\label{app:bench}

\subsection{LongBench Results}
\label{longbench}
We further evaluate RestoreKV on all 16 LongBench~\citep{longbench} tasks following the KVPress protocol. The evaluation comprises 3,750 examples with context lengths ranging from approximately 5K to 15K tokens (Qwen3 tokenizer). Table~\ref{tab:longbench_restorekv} reports results aggregated by task category.
RestoreKV improves the overall average across both backbones and both cache budgets, with larger gains at the tighter $r{=}0.0625$ budget.
At this ratio, the average score increases from $33.5$ to $37.7$ on
Llama-3.1-8B-Instruct and from $28.6$ to $32.1$ on Qwen3-8B.
These results extend the main-paper findings to a broader task suite, with the clearest gains under aggressive KV eviction.

\subsection{SCBench Results}
\label{scbench}
To evaluate generalization to substantially longer contexts, we additionally
consider nine SCBench~\cite{scbench} tasks with an average context length of approximately
104K tokens (Qwen3 tokenizer), following the evaluation protocol of KVzip
(Fig.~\ref{fig:scbench}). These tasks include Retr.KV,
Retr.Prefix-Suffix, and Retr.MultiHop for string retrieval; Code.RepoQA,
En.QA, and En.MultiChoice for semantic retrieval; and Math.Find,
ICL.ManyShot, and En.Sum for global-context understanding.

Focusing on the average scores, RestoreKV consistently improves KVzip
across all four KV budget ratios, raising the average from 29.4 to 29.7
at $r{=}0.4$, from 31.6 to 33.1 at $r{=}0.2$, from 30.0 to 32.4 at
$r{=}0.1$, and from 24.1 to 25.6 at $r{=}0.05$. Notably, RestoreKV is
trained only on contexts up to 15K tokens (Table~\ref{tab:traindata}),
so these gains reflect generalization to context lengths roughly an
order of magnitude beyond those seen during training. These results
suggest that the benefit of the restore cache is not limited to short
contexts and can extend to longer-context settings.

\section{Implementation and Reproducibility Details}
\label{app:implementation}

\begin{table}[t]\centering\small
\begin{tabular}{lrrr}
\toprule
\textbf{Source} & \textbf{Samples} & \textbf{Ctx (mean)} & \textbf{Ctx range} \\
\midrule
LongAlpaca   & 2{,}488 & 7{,}306 & 4{,}961--15{,}237 \\
PG-19        & 2{,}260 & 3{,}071 & 3{,}071--3{,}072  \\
Tulu-3 FLAN  & 1{,}500 & 898     & 512--9{,}283      \\
\midrule
\textbf{Total} & \textbf{6{,}248} & 4{,}236 & 512--15{,}237 \\
\bottomrule
\end{tabular}
\caption{RestoreKV training data statistics (context length in tokens).}
\label{tab:traindata}
\end{table}

\subsection{Training Data Construction}
\label{app:traindata}

We construct the training set from three sources
(Table~\ref{tab:traindata}).
\textit{(i) LongAlpaca self-study:}
We select 500 LongAlpaca~\cite{chen2024longlora} documents containing 2{,}048--16{,}000 tokens (Qwen3 tokenizer).
For each document, the teacher generates five questions spanning factual,
summarization, multi-hop, method, and comparison queries
($\leq 64$ tokens each), and answers them greedily over the same context
($\leq 512$ tokens). After filtering invalid or failed generations,
this process yields 2{,}488 examples.
\textit{(ii) PG-19 self-study:}
We split 50 PG-19~\cite{pg19} books into 3{,}072-token chunks, using up to 10 chunks per
book, and generate five question--answer pairs for each chunk, resulting in
2{,}260 examples after filtering.
\textit{(iii) Tulu-3 FLAN few-shot:}
We use 1{,}500 filtered examples from the FLAN v2 subset of
\texttt{allenai/tulu-3-sft-mixture} on Hugging Face~\citep{tulu,longpre2023flan},
paired with teacher-generated responses. The three sources form a training mixture of 6{,}248 examples. For computational efficiency, teacher responses are generated offline once for each target model and reused throughout
training, rather than regenerated at every optimization step.

\subsection{Training Setup Details}
\label{app:training_setup}
We keep the base model frozen and train only the $n{=}8$ restore-token embeddings and
LoRA adapters ($r_{\mathrm{LoRA}}{=}8$, $\alpha{=}16$, dropout 0) on the attention and MLP
projections ($\sim$16.5M parameters, 0.4\% of a 4B model). Optimization
uses AdamW ($\beta{=}(0.9,0.999)$, weight decay 0.01) with learning rate
$2\times10^{-4}$ for the restore tokens and LoRA, a cosine schedule with 50 warmup steps, gradient clipping at
norm 1.0, and a batch of one context per step for 5{,}000 steps. The objective is a symmetric KL distillation loss between the answer-token distributions of the restored-cache student using the final budget-matched cache and the frozen full-cache teacher. During
training the KV ratio is sampled from $\mathcal{U}(0.025, 0.25)$ to target
extreme-low budgets.

\begin{table}[t]
\centering
\small
\begin{tabular}{@{}l@{\hspace{1.4em}}l@{}}
\toprule
\multicolumn{2}{@{}l}{\textbf{Hardware}} \\
GPU        & NVIDIA RTX PRO 6000 Blackwell (96\,GB) \\
CPU        & Intel Xeon Gold 6530 \\
Memory     & 503\,GiB \\
% \addlinespace[5pt]
\midrule
\multicolumn{2}{@{}l}{\textbf{Software}} \\
OS         & Ubuntu 22.04.5 LTS \\
Framework  & PyTorch 2.8.0 (CUDA 12.8) \\
\bottomrule
\end{tabular}
\caption{\textbf{Experimental environment.}
Hardware and software configurations used for all experiments.}
\label{tab:env}
\end{table}

\begin{table}[t!]
\centering
\small
\setlength{\tabcolsep}{3pt}
\renewcommand{\arraystretch}{1.05}
\resizebox{\columnwidth}{!}{%
\begin{tabular}{@{}lcccc@{}}
\toprule
& \multicolumn{4}{c}{KV budget ratio} \\
\cmidrule(lr){2-5}
& $0.40$ & $0.20$ & $0.10$ & $0.05$ \\
\midrule
KVzip
& $93.46$
& $91.38$
& $80.08$
& $38.23$ \\

\textbf{RestoreKV}
& $\mathbf{94.25}\,{\scriptstyle\pm0.04}$
& $\mathbf{93.37}\,{\scriptstyle\pm0.11}$
& $\mathbf{88.21}\,{\scriptstyle\pm0.51}$
& $\mathbf{72.81}\,{\scriptstyle\pm0.57}$ \\
\bottomrule
\end{tabular}%
}
\caption{\textbf{Sensitivity to Training Seeds.}
RULER-4K accuracy on Qwen3-4B, reported as mean$\pm$std over
three training seeds.}
\label{tab:reproducibility}
\end{table}

\begin{table}[t!]
\centering
\small
\setlength{\tabcolsep}{4pt}
\renewcommand{\arraystretch}{1.08}
\begin{tabular}{@{}ll@{}}
\toprule
\textbf{Benchmark} & \textbf{Evaluation metric} \\
\midrule
RULER-4K
& Official string-matching score \\
QASPER
& Maximum token-F1 over reference annotations \\
QuALITY
& Answer-letter accuracy \\
LongHealth
& Answer-letter accuracy \\
LongBench
& Official per-task metric (F1, ROUGE-L, or accuracy) \\
SCBench
& Official task-specific evaluator \\
\bottomrule
\end{tabular}
\caption{\textbf{Evaluation metrics used for each benchmark.}}
\label{tab:eval_metrics}
\end{table}

\subsection{Baseline Adaptations (SnapKV and H$_2$O).}
\label{app:baseline_adaptations}

We adopt the query-agnostic, prefill-based adaptations of SnapKV~\cite{snapkv} and H$_2$O~\cite{h2o} used in the KVzip evaluation setup~\cite{kvzip}. In all cases, the context cache is scored and compressed before future queries are observed, and KV importance is evaluated at the individual KV-pair level under the same budget-allocation framework.

\paragraph{SnapKV}
We use the final $w{=}32$ context positions as the observation window and compute their attention over the context keys. For each KV pair, we average attention across the observation-window queries and apply max-pool smoothing with a kernel size of $7$. The trailing observation window is always retained. Because the observation window is drawn entirely from the context rather than from a future task query, the resulting cache is query-agnostic.

\paragraph{H$_2$O}
We use the prefill-based H$_2$O adaptation described in KVzip. For each KV pair, its importance is defined as the maximum attention it receives over all causal queries during context prefill. We use maximum rather than mean aggregation, following the finding in KVzip that maximum attention provides better compression performance. This baseline therefore derives importance from prefill self-attention, in contrast to the context-reconstruction attention used by KVzip.

\subsection{Experimental Environment}
\label{app:environment}

Our experimental environment is summarized in Table~\ref{tab:env}. For the comparison against KVPress Benchmark~\citep{kvpress} entries (RULER and LongBench) we implemented our
method within the official KVPress codebase and ran those evaluations
in its environment, ensuring a fair comparison
against published baselines.

\subsection{Sensitivity to Training Seeds}
\label{app:training_robustness}

Inference in RestoreKV is deterministic: importance scoring involves no
sampling, and decoding is greedy.
The remaining stochasticity comes from training, including random
initialization and retention-ratio sampling.
All main results report the seed-0 run.
To assess training robustness, we additionally train RestoreKV with two
different seeds and evaluate the three runs on RULER-4K.
As shown in Table~\ref{tab:reproducibility}, the standard deviation remains
below $0.57$ across all cache ratios.
At $r{=}0.05$, this variation is approximately $60\times$ smaller than the
$34.6$-point mean improvement over KVzip.

\subsection{Evaluation Metrics}
\label{app:eval_metrics}

We follow the official evaluation metric of each benchmark to ensure
consistency with its standard evaluation protocol and comparability with
prior work. RULER and LongBench are evaluated following the KVPress
protocol, while SCBench uses its official task-specific evaluators.
QASPER uses the maximum token-level F1 over reference annotations, whereas
QuALITY and LongHealth use answer-letter accuracy.
Table~\ref{tab:eval_metrics} summarizes the metrics.

\end{document}